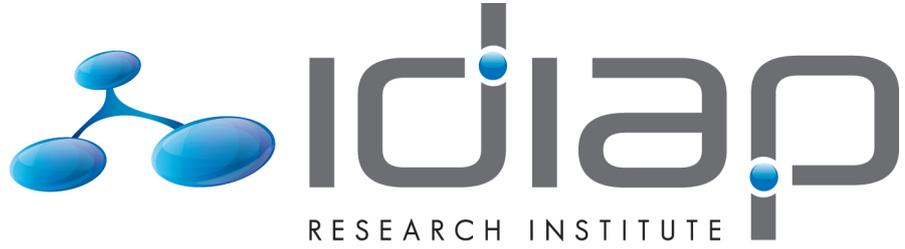

# RECURRENT CONVOLUTIONAL NEURAL NETWORKS FOR SCENE PARSING

Pedro H. O. Pinheiro  Ronan Collobert

# Recurrent Convolutional Neural Networks for Scene Parsing

Pedro H. O. Pinheiro and Ronan Collobert

Idiap Research Institute, Rue Marconi 19, 1920 Martigny, Switzerland
pedro.pinheiro@idiap.ch,ronan.collobert@idiap.ch

**Abstract.** Scene parsing is a technique that consist on giving a label to all pixels in an image according to the class they belong to. To ensure a good visual coherence and a high class accuracy, it is essential for a scene parser to capture image long range dependencies. In a *feed-forward* architecture, this can be simply achieved by considering a sufficiently large input context patch, around each pixel to be labeled. We propose an approach consisting of a recurrent convolutional neural network which allows us to consider a large input context, while limiting the capacity of the model. Contrary to most standard approaches, our method does not rely on any segmentation methods, nor any task-specific features. The system is trained in an end-to-end manner over raw pixels, and models complex spatial dependencies with low inference cost. As the context size increases with the built-in recurrence, the system identifies and corrects its own errors. Our approach yields state-of-the-art performance on both the Stanford Background Dataset and the SIFT Flow Dataset, while remaining very fast at test time.

## 1  Introduction

In the computer vision field, *scene parsing* is the task of fully labeling an image pixel-by-pixel with the class of the object the pixel belongs to. This task is very challenging, as it implies solving a detection, a segmentation and a multi-label recognition problem all in one.

The parsing problem is most commonly addressed with some kind of *local* classifier constrained in its predictions with a graphical model (*e.g.* Conditional Random Fields, Markov Random Fields), such that a *global* decision is made. These approaches usually consist of segmenting the image into superpixels or segment regions to assure a visible consistency of the labeling and also to take into account similarities between neighbor segments, giving a high level understanding of the overall structure of the image. Each segment contains a series of input features describing it and contextual features describing spatial relation between the label of neighbor segments. These models are then trained to maximize the likelihood of correct classification given the features [1–7]. The main limitation of scene parsing approaches based on graphical models is the computational cost at test time, limiting the model to simple contextual features.

In this work, we consider a *neural network-based feed-forward approach* which can take into account long range dependencies on the image while controlling the capacity of the network, achieving state-of-the-art accuracy while keeping the computational cost low at test time, thanks to the complete feed-forward design. Our method relies on a recurrent architecture for convolutional neural networks: a series of networks sharing the same set of parameters. Each instance consider as input both an RGB image and the classification attempt of the previous instance of the network. The network learns itself how to smooth its own predicted labels, improving the estimation as the number of instances increases.

Compared to graphical models approaches relying on image segmentation, our system has several advantages: (i) it does not require any engineered features, since deep learning architectures train (hopefully) adequate discriminative filters in an end-to-end manner, (ii) the prediction phase does not rely in any label space searching, since it requires only the *forward evaluation of a function*.

In the following, Section 2 briefly presents related works, Section 3 describe the proposed strategy, Section 4 presents the results of our experiments in two standard datasets: the Stanford Background Dataset (8 classes) and the SIFT Flow Dataset (33 classes) and compare the performance with other systems. The paper is finished with a conclusion and discussion on Section 5.

## 2   Related Work

In a preliminary work, [8] proposed an innovative approach to scene parsing without the use of any graphical model. The authors propose a solution based on deep convolutional networks relying on a *supervised* greedy learning strategy. These network architectures can be fed with raw pixels and are able to capture texture, shape and contextual information.

[6] also considered the use of deep learning techniques to deal with scene parsing. Unlike us, the authors consider off-the-shelf features of segments obtained from the scenes. They then use a network for recursively merging different segments and give them a semantic category label. Our recurrent architecture differs from theirs in the sense that we use it to parse the scene with a smoother class annotation.

More recently, [9] also consider the use of convolutional networks, extracting features densely from a multiscale pyramid of images. This solution yields satisfactory results for the categorization of the pixels, but poor visual coherence.

The authors propose three different over-segmentation approaches to produce the final labeling with improved accuracy and better visual coherence: (i) the scene is segmented in superpixels and a single class is assigned to each of the superpixels, (ii) a conditional random field is defined over a set of superpixels to model joint probabilities between them and correct aberrant pixel classification (such as "road" pixel surrounded by "sky") and (iii) the selection of a subset of tree nodes that maximize the average "purity" of the class distribution, hence maximizing the overall likelihood that each segment will contain a single object. The superpixel-based approach was however order of magnitude faster than the two other approaches, with slightly lower performance in accuracy. In contrast, our approach is simpler and completely feed-forward, as it does not require any image segmentation technique, nor the handling of a multiscale pyramid of input images.

**Table 1.** Comparison between different methods for scene parsing. The advantage of our proposed method consists on the simplicity of inference, not relying on any task-specific feature extraction nor segmentation method.

| METHOD | TASK-SPECIFIC FEATURES |
|---|---|
| GOULD ET AL., 2009 [1] | 17-DIMENSIONAL COLOR AND TEXTURE FEATURES, 9 GRID LOCATIONS AROUND THE PIXEL AND THE IMAGE ROW, REGION SEGMENTATION. |
| TIGHE & LAZEBNIK, 2010 [3] | GLOBAL, SHAPE, LOCATION, TEXTURE/SIFT, COLOR, APPEARANCE, MRF. |
| MUNOZ ET AL., 2010 [15] | GIST, PYRAMID HISTOGRAM OF ORIENTED GRADIENTS, COLOR HISTOGRAM CIELAB, RELATIVE RELOCATION, HIERARCHICAL REGION REPRESENTATION. |
| KUMAR & KOLLER, 2010 [2] | COLOR, TEXTURE, SHAPE, PERCENTAGE PIXELS ABOVE HORIZONTAL, REGION-BASED SEGMENTATION. |
| SOCHER ET AL., 2012 [6] | SAME AS [1]. |
| LEMPITSKY ET AL., 2011 [7] | HISTOGRAM OF VISUAL SIFT, HISTOGRAM OF RGB, HISTOGRAM OF LOCATIONS, "CONTOUR SHAPE" DESCRIPTOR. |
| FARABET ET AL., 2013 [9] | LAPLACIAN PYRAMID, SUPERPIXELS/CRF/TREE SEGMENTATION. |
| OUR RECURRENT CNN | RAW PIXELS |

As in [9], [10] proposed a similar multiscale convolutional architecture. In their approach, the authors smooth out the predicted labels with pairwise class filters.

Compared to existing approaches, our method does not rely on any task-specific feature (see Table 1). Our scene parsing system is able to extract relevant contextual information from raw pixels.

## 3  Systems Description

We formally introduce convolutional neural networks (CNNs) in Section 3.1. In Section 3.2 we discuss how to capture long range dependencies with these type of models, while keeping a tight control on the capacity. Section 3.3 introduces our recurrent network approach for scene parsing. Finally, in Section 3.4, we show how to infer the full scene label. Because of pooling operations in a convolutional neural network architecture, the predicted output label "planes" have a lower resolution than the input image. Here we introduce a way to obtain the full resolution in an efficient way.

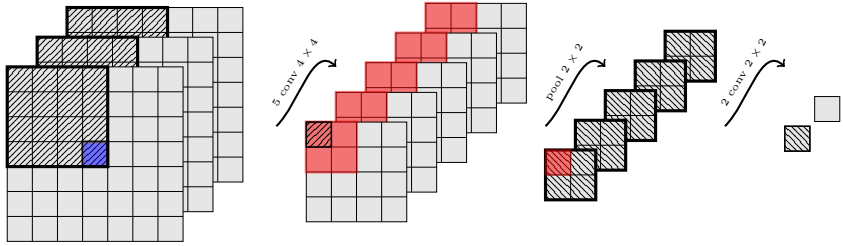

**Fig. 1.** A simple convolutional network. Given an image patch providing a context around a pixel to classify (here blue), a series of convolutions and pooling operations (filters slid through input planes) are applied (here 5 4 × 4 convolutions, followed by a 2 × 2 pooling, followed by 2 2 × 2 convolutions. Each 1 × 1 output plane is interpreted as a score for a given class.

### 3.1  Convolutional Neural Networks for Scene Parsing

Convolutional neural networks [11] are a natural extension of neural networks for treating images. Their architecture, vaguely inspired by the biological visual system, possesses two key properties that make them extremely useful for image applications: spatially shared weights and spatial pooling. These kind of networks learn features that are shift-invariant, *i.e.*, filters that are useful across the entire image (due to the fact that image statistics are stationary). The pooling layers are responsible for reducing the sensitivity of the output to slight input shift and distortions. This type of neural network has proven to be very efficient in many vision applications, such as object recognition and segmentation ([12, 13]).

A typical convolutional network is composed of multiple stages, as shown on Figure 1. The output of each stage is made of a set of two dimensional arrays called feature maps. Each feature map is the outcome of one convolutional (or pooling) filter applied over the full image. A non-linear squashing function (such as a hyperbolic tangent) always follows a pooling layer.

In the context of scene parsing we consider a set of images indexed by $I_k$, and we are interested in finding the label of each pixel at location $(i, j)$, for every image $k$. To that matter, the network is fed with a squared *context patch* $I_{i,j,k}$ surrounding the pixel at location $(i, j)$ in the $k$-th image. It can be shown (see Figure 1) that the output plane size $sz_m$ of the $m^{th}$ layer is computed as:

$$sz_m = \frac{sz_{m-1} - kW_m}{dW_m} + 1 , \qquad (1)$$

where $sz_0$ is the input patch size, $kW_m$ is the size of the convolution (or pooling) kernels in the $m^{th}$ layer, and $dW_m$ is the pixel step size used to slide the convo-lution (or pooling) kernels over the input planes.[1] Given a network architecture and an input image, one can compute the output image size by successively applying Equation 1 on each layer of the network. During the training phase, the size of the input patch $I_{i,j,k}$ is chosen carefully such that the output layers produces $1 \times 1$ planes, which are then interpreted as scores for each class of interest.

Adopting the same notation as [9], the output of a network $f$ with $M$ stages and trainable parameters $(\mathbf{W}, \mathbf{b})$, given an input patch $I_{i,j,k}$, can be formally written as:

$$f(I_{i,j,k}; (\mathbf{W}, \mathbf{b})) = \mathbf{W}_M \mathbf{H}_{M-1}, \qquad (2)$$

with the output of the $m^{th}$ hidden layer computed as:

$$\mathbf{H}_m = \tanh(\text{pool}(\mathbf{W}_m \mathbf{H}_{m-1} + \mathbf{b}_m)), \qquad (3)$$

for $m = \{1, ..., M\}$ and denoting $\mathbf{H}_0 = I_{i,j,k}$. $\mathbf{b}_m$ is the bias vector of layer $m$ and $\mathbf{W}_m$ is the Toeplitz matrix of connection between layer $m-1$ and layer $m$. The pool$(\cdot)$ function is the max-pooling operator and $\tanh(\cdot)$ is the point-wise hyperbolic tangent function applied at each point of the feature map.

In order to train the network by maximizing a likelihood, the network scores $f_c(I_{i,j,k}; (\mathbf{W}, \mathbf{b}))$ (for each class of interest $c \in \{1, ..., N\}$) are transformed into conditional probabilities, by applying a *softmax* function:

$$p(c|I_{i,j,k}; (\mathbf{W}, \mathbf{b})) = \frac{e^{f_c(I_{i,j,k}; (\mathbf{W}, \mathbf{b}))}}{\sum\limits_{d \in \{1, ..., N\}} e^{f_d(I_{i,j,k}; (\mathbf{W}, \mathbf{b}))}} \qquad (4)$$

The parameters $(\mathbf{W}, \mathbf{b})$ are learned in an end-to-end supervised way, by minimizing the negative log-likelihood over the training set:

$$L(\mathbf{W}, \mathbf{b}) = -\sum_{I_{(i,j,k)}} \ln p(l_{i,j,k}|I_{i,j,k}; (\mathbf{W}, \mathbf{b})), \qquad (5)$$

where $l_{i,j,k}$ is the correct pixel label class, at position $(i,j)$ in image $I_k$. The minimization was achieved with the Stochastic Gradient Descent (SGD) algorithm, with a fixed learning rate $\lambda$:

$$\mathbf{W} \longleftarrow \mathbf{W} - \lambda \frac{\partial L}{\partial \mathbf{W}} \;;\; \mathbf{b} \longleftarrow \mathbf{b} - \lambda \frac{\partial L}{\partial \mathbf{b}}. \qquad (6)$$

### 3.2  Long Range Dependencies with Convolutional Networks

Existing successful scene parsing systems leverage long range image dependencies in some way. The most common approach is to add a kind of graphical model (*e.g.* a conditional random field) over local decisions, such that a certain global coherence is maintained. An obvious way to capture long range dependencies would be to consider large input patches when labeling a pixel. Unfortunately, this approach might face generalization issues, as considering large context often implies considering large models (*i.e.* large capacity).

---

[1] Most people use $dW = 1$ for convolutional layers, and $dW = kW$ for pooling layers.

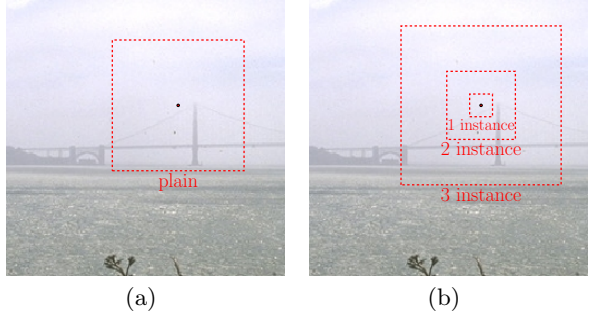

(a)                    (b)

**Fig. 2.** Context input path of "plain" (a) and recurrent (b) architecture. The size of the contextual input patch (b) increases as the number of instances in the recurrent convolutional network increases. The capacity of the model remains the same, since the parameters over all instances are shared.

In Table 2, we review possible ways to control the capacity of a convolutional neural network, assuming a large input context. In a "plain" architecture (as described in Section 3.1), one can easily control the capacity by increasing the filter sizes in pooling layers, reducing the overall number of parameters in the network. Unfortunately, performing large poolings decreases the network label output resolution (*e.g.*, if one performs a 1/8 pooling, the label output plane size will be about $1/8^{th}$ of the input image size). One can overcome this problem (as shown in Section 3.4), but at the cost of a slow inference process.

Instead, Farabet et al., 2013 [9] considered the use of a *multiscale* convolutional network. Large contexts are integrated into local decisions while making the model still manageable in terms of parameters/dimensionality. Label coherence is then increased by leveraging superpixels.

Another way to consider a large input context size while controlling the capacity of the model is to make the network recurrent: the output of the model is fed back to the input of another *instance* of the same network, which *shares* the same parameters (see Figure 3). Given Equation 1, we have

$$sz_{m-1} = dW_m \times sz_m + (kW_m - dW_m).$$

Thus, the required context to label one pixel (*i.e.* if the network output size is $1\times 1$), increases with the number of network instances (see an example in Figure 2). However, the capacity of the system remains constant, since the parameters of each network instance are simply shared. We will now detail our recurrent network approach.

**Table 2.** Long range dependencies integration in CNN-based scene parsing systems. Ways to control capacity and speed of each architecture is reported.

| Means | Capacity control | Speed |
|---|---|---|
| LOCAL CLASSIFIER + GRAPHICAL MODEL | – | SLOW |
| MULTISCALE | SCALE DOWN INPUT IMAGE | FAST |
| LARGE INPUT PATCHES | INCREASE POOLING | SLOW |
|  | RECURRENT ARCHITECTURE | FAST |

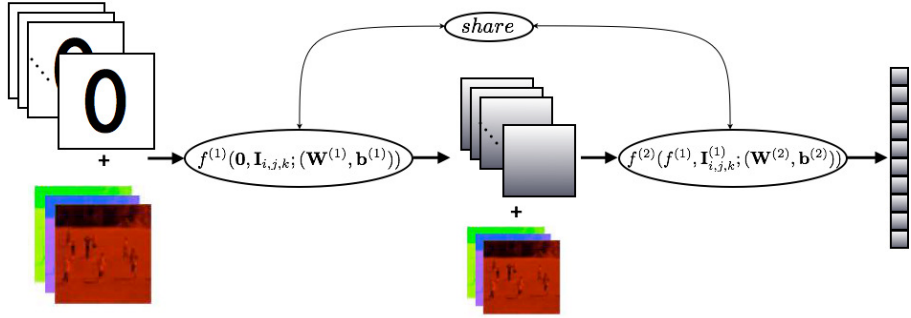

**Fig. 3.** Recurrent network architecture with two instance of $f$. The first instance $f^{(1)}$ of the recurrent architecture is fed with a RGB patch and an empty feature map. The output of the first network is coupled with the scaled RGB patch and fed to the same network (shared parameters ($\mathbf{W}$ and $\mathbf{b}$)).

### 3.3  Recurrent Network Approach

The recurrent architecture (see Figure 3) consists of $P$ instances of the "plain" convolutional network $f(\cdot)$, each of them with *identical* (shared) trainable parameters ($\mathbf{W}, \mathbf{b}$). Each instance $f^p$ ($1 \leq p \leq P$) is fed with an input "image" $\mathbf{F}^p$ of $N+3$ features maps

$$\mathbf{F}^p = [f^{p-1}(I_{i,j,k}^{p-1};(\mathbf{W},\mathbf{b})), I_{i,j,k}^p],\ \mathbf{F}^1 = [\mathbf{0}, I_{i,j,k}],$$

which are the output label planes of the previous instance, and the scaled[2] version of the raw RGB squared patch surrounding the pixel at location $(i,j)$ of the training image $k$. To make the number of inputs equal in all instances of the network, the input "image" of the first instance is simply the patch of raw pixel coupled with $N$ 0 feature maps. This guarantees the end-to-end characteristics of the system as well as its fast inference during test.

---

[2] $I_{i,j,k}^{p-1}$ is scaled so that it has the width/height as $f^{p-1}$.

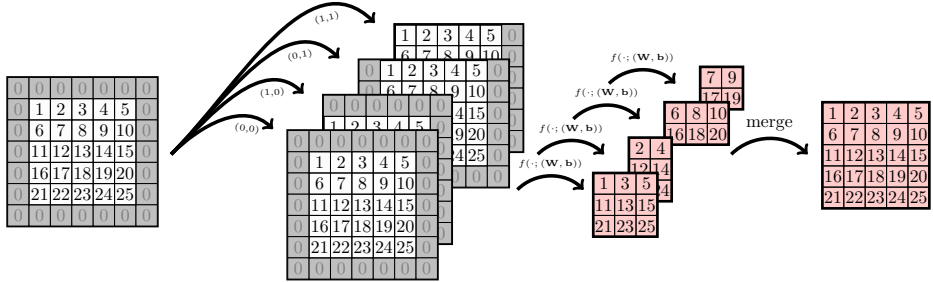

**Fig. 4.** Convolutional neural networks output downscaled label planes (compared to the input image) due to pooling layers. To alleviate this problem, one can feed several shifted version of the input image (here represented by pixels 1..25) in the X and Y axis. In this example the network is assumed to have a single 2 × 2 pooling layer. Downscaled predicted label planes (here in red) are then merged to get back the full resolution label plane in an efficient manner. Note that pixels represented by 0 are adequate padding.

As in the "plain" network, the size of the patch during training is chosen such that the output layers produces $1 \times 1$ planes, which are then interpreted as scores for each class of interest.

### 3.4 Scene Inference

Given a test image $I$, the pixel at location $(i, j)$ of image $k$ is labeled with the *argmax* of the network predictions:

$$\hat{l}_{i,j,k} = \arg\max_{c \in \text{classes}} p(c | I_{i,j,k}; (\mathbf{W}, \mathbf{b})), \tag{7}$$

considering the context patch $I_{i,j,k}$. Note that this might imply adding padding when inferring label of pixels close to the image border. In practice, simply extracting patches $I_{i,j,k}$ and then feeding them through the network for all pixels of a test image is computationally very inefficient. Instead, it is a useful practice to feed the full test image (also properly padded) to the convolutional network: *applying one convolution to a large image is much faster than applying the same convolution many times to small patches*. When fed with the full input image, the network will output a plane of label scores. Unfortunately, following Equation 1, the plane size is smaller than the input image size: this is mainly due to pooling layers, but also to border effects when applying the convolution. E.g., if the network includes two $2 \times 2$ pooling layers, only 1 every 4 pixels of the input image will be labeled. Most convolutional network users (see ([9])) upscale the label plane to the input image size.

In fact, it is possible to compute efficiently the label plane with a fine resolution by feeding to the network several versions of the input image, shifted on the $X$ and $Y$ axis. We show an example on Figure 4, for a network which would have only one $2 \times 2$ pooling layer, and one output plane: low resolution label planes (coming out of the network for the input image shifted by $(0,0)$, $(0,1)$, $(1,0)$ and $(1,1)$ pixels) are merged to form the high resolution label plane. If one has several pooling layers which downscale more the input image (e.g. by a factor of 8), one has to do more forwards (e.g. $8 \times 8 = 64$), which is still much faster than forwarding patches at each location of the test image. We will see in Section 4.3 that having a finer label resolution can increase the classification performance.

**Table 3.** Pixel and averaged per class accuracy and the computing time of other methods and our proposed approaches on the Stanford Background Dataset.

| METHOD | PER PIXEL ACCURACY | AVG PER CLASS ACCURACY | COMPUTE TIME (S) |
|---|---|---|---|
| Gould et al., 2009 [1] | 76.4% | - | 10 to 600 |
| Tighe & Lazebnik, 2010 [3] | 77.5% | - | 10 to 300 |
| Munoz et al., 2010 [15][3] | 76.9% | 66.2% | 12 |
| Kumar & Koller, 2010 [2] | 79.4% | - | < 600 |
| Socher et al., 2012 [6] | 78.1% | - | ? |
| Lempitsky et al., 2011 [7] | 81.9% | 72.4% | > 60 |
| Farabet et al., 2013 [9][4] | 78.8% | 72.4% | 0.6 |
| Farabet et al., 2013 [9][5] | 81.4% | 76.0% | 60.5 |
| Plain CNN | 79.5% | 69.5% | 15 |
| $CNN_1$ | 67.9% | 58.0% | 0.2 |
| $rCNN_1$ (2 instances) | 79.4% | 69.5% | 2.6 |
| $CNN_2$ | 15.3% | 14.7% | 0.06 |
| $rCNN_2$ (2 instances) | 76.2% | 67.2% | 1.1 |
| $rCNN_2$ 1/2 resolution (3 instances) | 79.8% | 69.3% | 2.15 |
| $rCNN_2$ full resolution (3 instances) | 80.2% | 69.9% | 10.7 |

## 4   Experiments

We tested our proposed method on two different datasets for scene parsing: the Stanford Background [1] and the SIFT Flow Dataset [5]. The Stanford dataset has 715 images from rural and urban scenes composed of 8 classes. The scenes have approximately $320 \times 240$ pixels. As in [1], we performed a 5-fold cross-validation with the dataset randomly split into 572 training images and 143 test images in each fold. The SIFT Flow is a larger dataset composed of 2688 images of $256 \times 256$ pixels and 33 semantic labels. All the algorithms and experiments were implemented in Torch7 [14].

Each image on the training set was properly padded and normalized such that they have zero mean and unit variance. All networks were trained by sampling patches surrounding randomly chosen pixel from randomly chosen images from the training set of images. Contrary to [9] (i) we did not consider any extra distortion on the images[6], and (ii) we did not sample training patches according to balanced class frequencies.

---
[3] Unpublished improved results have been recently reported by the authors.
[4] Multiscale CNN + superpixels.
[5] Multiscale CNN + CRF.

**Table 4.** Pixel and averaged per class accuracy of other methods and our proposed approaches on the SIFT Flow Dataset.

| Method | Per Pixel Accuracy | Avg Per Class Accuracy |
|---|---|---|
| Liu et al., 2009 [5] | 74.75% | - |
| Tighe & Lazebnik, 2010 [3] | 76.9% | 29.4% |
| Farabet et al., 2013 [9] | 78.5% | 29.6% |
| Plain | 76.5% | 31.0% |
| $CNN_1$ | 51.8% | 17.4% |
| $rCNN_1$ | 76.2% | 29.2% |
| $rCNN_2$ (2 instances) | 65.5% | 20.8% |
| $rCNN_2$ (3 instances) | 77.7% | 29.8% |

We considered two different accuracy measures to compare the performance of our proposed methods with other approaches. The first one is the accuracy per pixel of test images. This measure is simply the ratio of correct classified pixels of all images in the test set. However, in scene parsing (especially in datasets with large number of classes), classes which are much more frequent than others (*e.g.* the "sky" class is much more frequent than "moon") have more impact on this measure. Recent papers also consider the averaged per class accuracy on the test set (all classes have the same weight in the measure). Note that as mentioned above, we did not train with balanced class frequencies, which would have optimized this second measure.

We consider both a "plain" architecture with a large patch and strong number of pooling and a recurrent architecture with two and three unfolds in time. Table 3 compares the performance of our architecture with related works on the Stanford Background Dataset and Table 4 compares the performance on the SIFT Flow Dataset. In the following, we provide additional technical details for each architecture used.

### 4.1  Plain Network

The first "plain" network was trained with $133 \times 133$ input patches. The network was composed of a $6 \times 6$ convolution with $nhu_1$ output planes, followed by a $8 \times 8$ pooling layer, a $\tanh(\cdot)$ non-linearity, another $3 \times 3$ convolutional layer with $nhu_2$ output planes, a $2 \times 2$ pooling layer, a $\tanh(\cdot)$ non-linearity, and a final $7 \times 7$ convolution to produce label scores. The hidden units were chosen to be $nhu_1 = 25$ and $nhu_2 = 50$ for the Stanford dataset, and $nhu_1 = 50$ and $nhu_2 = 50$ for the SIFT Flow dataset.

---

[6] Which is known to improve the generalization accuracy by few extra percents.

### 4.2 Recurrent Architectures

We consider two different recurrent convolutional network architectures.

The first architecture, $rCNN_1$, is composed of two consecutive instances of the convolutional network $CNN_1$, sharing parameters (as in Figure 3). $CNN_1$ is composed of a $8 \times 8$ convolution with 25 output planes, followed by a $2 \times 2$ pooling layer, a $\tanh(\cdot)$ non-linearity, another $8 \times 8$ convolutional layer with 100 output planes, a $2 \times 2$ pooling layer, a $\tanh(\cdot)$ non-linearity, and a final $1 \times 1$ convolution to produce $N$ label scores.

$rCNN_1$ is trained by considering the two network instances simultaneously. For each training example we randomly choose to perform a "forward" and "backward" pass through one or two instances of the network. This training approach allows the network to learn to correct its own mistakes (made by the first network instance). As mentioned in Section 3.2, the input context patch size depends directly on the number of network instances in the recurrent architecture. In the case of $rCNN_1$, the patch size is of $25 \times 25$ when considering one instance and $121 \times 121$ when considering two network instances.

The second recurrent convolutional neural network, $rCNN_2$, is composed of three instances of the convolutional network $CNN_2$, with shared parameters. Each instance of $CNN_2$ is composed of a $8 \times 8$ convolution with 25 output planes, followed by a $2 \times 2$ pooling layer, a $\tanh(\cdot)$ non-linearity, another $8 \times 8$ convolution with 50 planes and a final $1 \times 1$ convolution which outputs the $N$ label planes.

In $rCNN_2$, the first two instances are trained simultaneously[7] (as in Figure 3) through SGD, with input patch of size 67. Then, after the system with two instances are trained, a third instance of the network is considered (still with the parameters shared with the others instances) so that it is able to correct itself from the previous labeling. The input patch is of size 155 in this latter case.

Figure 5 illustrates inference of the recurrent network with two and three instances. The network learns itself how to correct its own label prediction.

In all cases, the learning rate in Equation 6 was equal to $10^{-4}$. All hyper-parameters were tuned with a 10% hold-out for validation.

### 4.3 Compute Time and Scene Inference

In Table 5, we analyze the tradeoff between compute time and test accuracy, by running several experiments with different output resolutions for recurrent network $rCNN_2$ (see Section 3.4 and Figure 4). Labeling about $1/4^{th}$ of the pixels seemed to be enough to lead to near state-of-the-art performance, while keeping a very fast inference time.

---

[7] Considering only one instance in this case produces a very small context input patch.

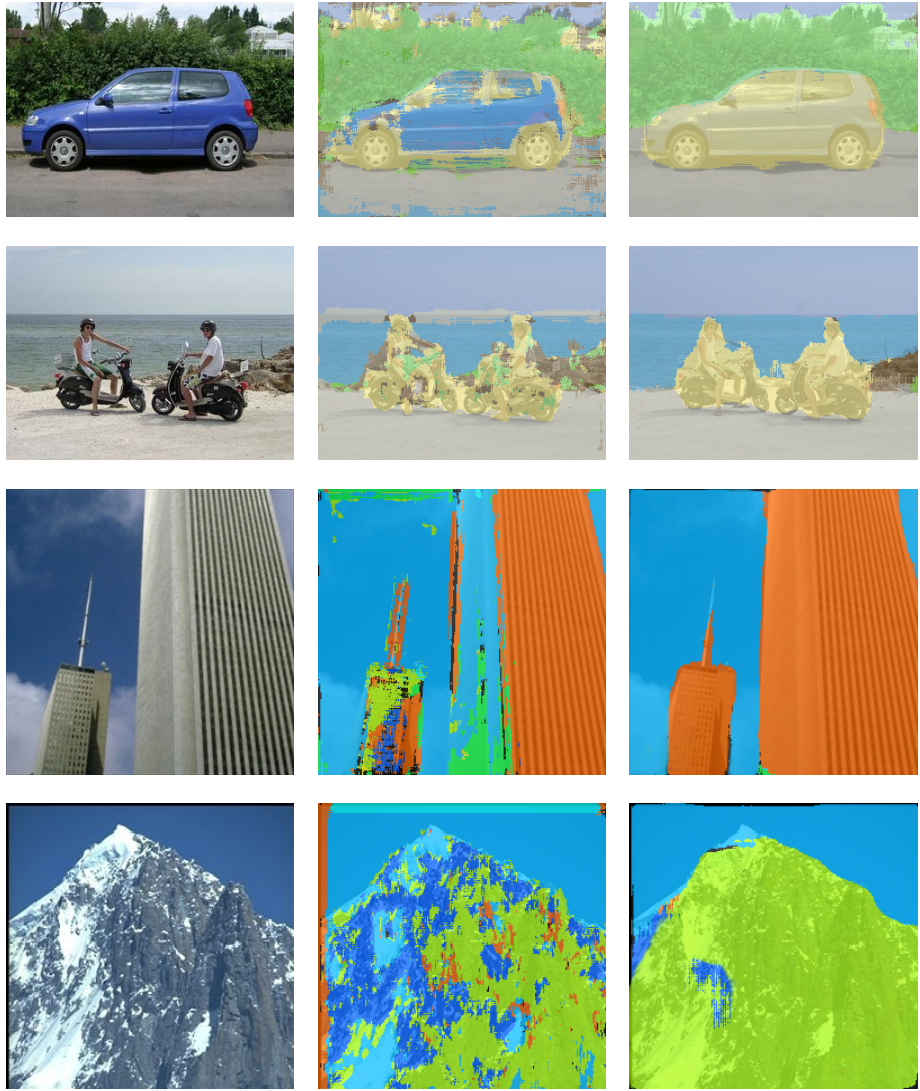

**Fig. 5.** Recurrent scene parser over original image (a) segments the image as shown in (b). Due to its recurrent nature, it can be fed again with its own prediction (b) and the original image (a) which leads to (c): most mistakes are corrected.

## 5   Conclusion

This paper presents a feed-forward approach for scene parsing, based on supervised "deep" learning strategies which model in a rather simple way non-local class dependencies in a scene from raw pixels. We demonstrate that the problem of scene parsing can be faced without the need of any expensive graphical model nor segmentation tree technique to ensure labeling. The scene labeling is inferred simply by forward evaluation of a function applied to a RGB image.

**Table 5.** Computation time and performance in pixel accuracy for the recurrent convolutional network rCNN$_2$ with different label resolution on the Stanford dataset. Our algorithms were ran on a 4-core Intel i7.

| Output Resolution | Compute time per image | Pixel accuracy |
|---|---|---|
| 1/8 | 0.20s | 78.4% |
| 1/4 | 0.70s | 79.3% |
| 1/2 | 2.15s | 79.8% |
| 1/1 | 10.68s | 80.2% |

In terms of accuracy, our system achieved state-of-the-art results on both Stanford Background and SIFT Flow dataset. Future work includes investigation of unsupervised or semi-supervised pre-training of the models, as well as application to larger datasets such as the Barcelona dataset.